%% file: main.tex
\documentclass{article}
\usepackage{ijcai20}

\usepackage{times}
\usepackage{soul}
\usepackage{url}
\usepackage[hidelinks]{hyperref}
\usepackage[utf8]{inputenc}
\usepackage[small]{caption}
\usepackage{graphicx}
\usepackage{amsmath}
\usepackage{amssymb}
\usepackage{booktabs}
\usepackage{algorithm}
\usepackage{algorithmic}
\usepackage{multirow}
\usepackage{makecell}
\usepackage{xcolor}
\usepackage{stackengine}

\title{Generating Person Images with Appearance-aware Pose Stylizer}
\author{
Siyu Huang$^1$
\and
Haoyi Xiong$^1$
\and
Zhi-Qi Cheng$^2$
\and
Qingzhong Wang$^3$ 
\and \\
Xingran Zhou$^4$ 
\and 
Bihan Wen$^5$
\and
Jun Huan$^6$
\And
Dejing Dou$^1$
\affiliations
$^1$Baidu Research ~~~
$^2$Carnegie Mellon University ~~~
$^3$City University of Hong Kong\\
$^4$Zhejiang University ~~~
$^5$Nanyang Technological University ~~~
$^6$Styling AI
\emails
\{huangsiyu, xionghaoyi, doudejing\}@baidu.com,
zhiqic@cs.cmu.edu,
qingzwang2-c@my.cityu.edu.hk,
xingranzh@zju.edu.cn,
bihan.wen@ntu.edu.sg,
lukehuan@shenshangtech.com
}

\begin{document}

\maketitle

\begin{abstract}
Generation of high-quality person images is challenging, due to the sophisticated entanglements among image factors, e.g., appearance, pose, foreground, background, local details, global structures, etc. In this paper, we present a novel end-to-end framework to generate realistic person images based on given person poses and appearances. The core of our framework is a novel generator called Appearance-aware Pose Stylizer (APS) which generates human images by coupling the target pose with the conditioned person appearance progressively. The framework is highly flexible and controllable by effectively decoupling various complex person image factors in the encoding phase, followed by re-coupling them in the decoding phase. In addition, we present a new normalization method named adaptive patch normalization, which enables region-specific normalization and shows a good performance when adopted in person image generation model. Experiments on two benchmark datasets show that our method is capable of generating visually appealing and realistic-looking results using arbitrary image and pose inputs. 
\end{abstract}

\section{Introduction}
\begin{figure}[t]
\centering
\includegraphics[width=1\linewidth]{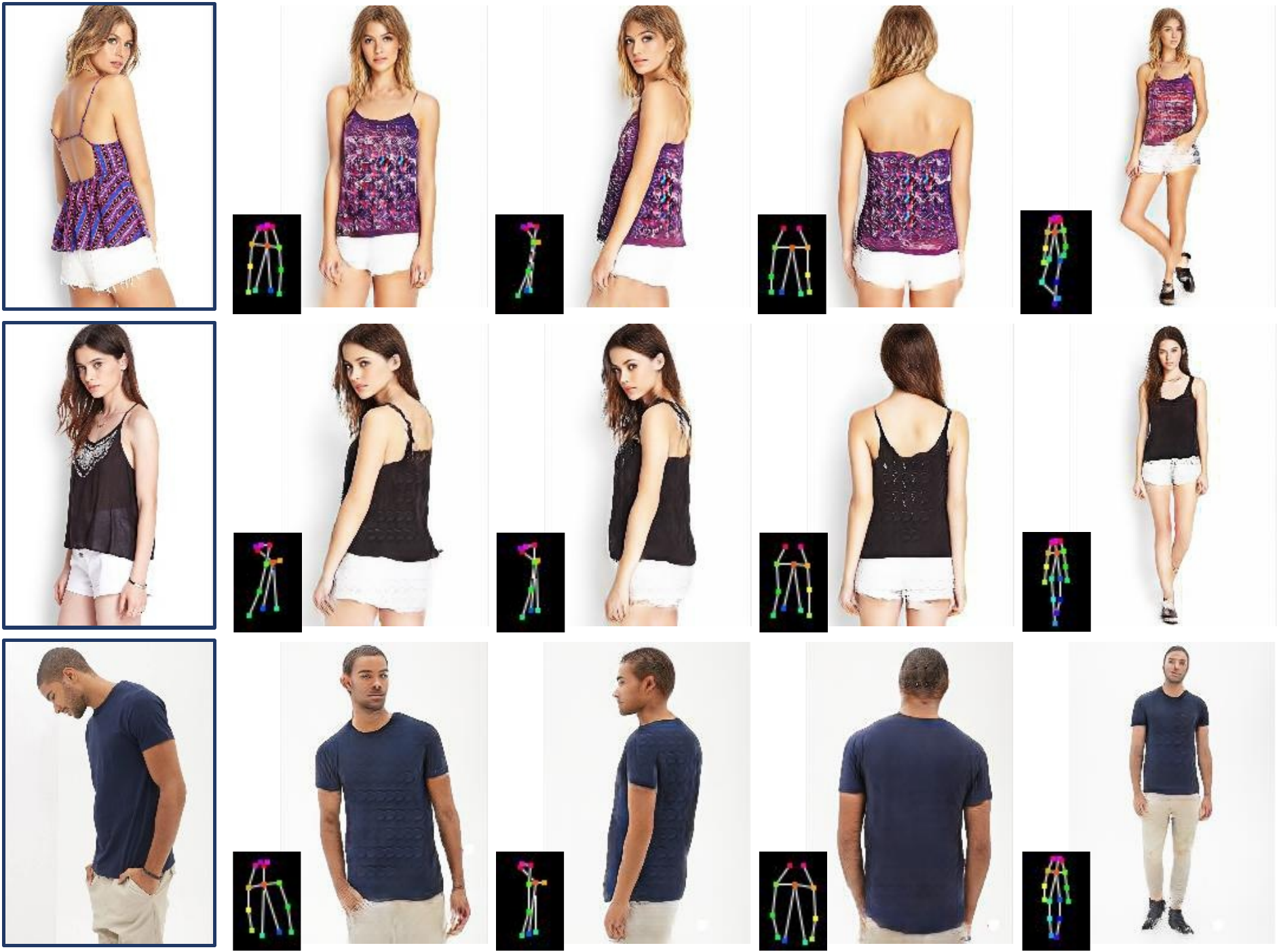}
\caption{Examples of pose-guided person generation. The conditioned source images are shown at left. The target postures and the person images generated by our method are shown at right. Our method shows realistic and appealing results.}
\label{fig:fig1}
\end{figure}

Generating realistic-looking human images is of great value in many tasks such as surveillance data augmentation \cite{zheng2017unlabeled} and video forecasting \cite{walker2017pose,wang2018video}. In this work\footnote{Code is available at \textit{\url{https://github.com/siyuhuang/PoseStylizer}}}, we focus on the pose-guided person image generation \cite{pg2} which aims to transfer person images from one pose to other poses. The generated person is expected to accord with the conditioned pose structure as well as preserving the appearance details of the source person. Fig. \ref{fig:fig1} provides some pose transfer results generated by our proposed framework as examples. The pose-guided person image generation is very challenging due to the following aspects: (1) The distributions of clothes, body appearance, backgrounds, and poses vary largely between human images; (2) One person of different poses may have very different visual features; (3) The generative model usually needs to infer the appearance details of body parts which are unobserved in input images.

To address the above challenges, it is essential to decouple the complex entanglements, such as the interplay between appearance and pose \cite{vunet}, in person generation procedure. Towards that goal \cite{disentangle} introduced a learning-based pipeline to disentangle and encode three factors: image foreground, background, and pose into separated representations and then decode them back to a person image. Although the three factors are successfully decoupled by the encoder, the representations from three factors are simply concatenated into latent codes before they are fed into the decoder, resulting in a lack of controllability and interpretability in the decoding phase. We note that such a mode decoupling is significant for the decoding phase. From the application perspective, the users prefer a more controllable image editing process brought by flexible input modes. From the engineering perspective, the experts would lean to a generator with more interpretability, such that distinct feedback can be collected from disentangled modes to better guide model design, tuning, and optimization. 

\begin{figure}[t]
\centering
\includegraphics[width=1\linewidth]{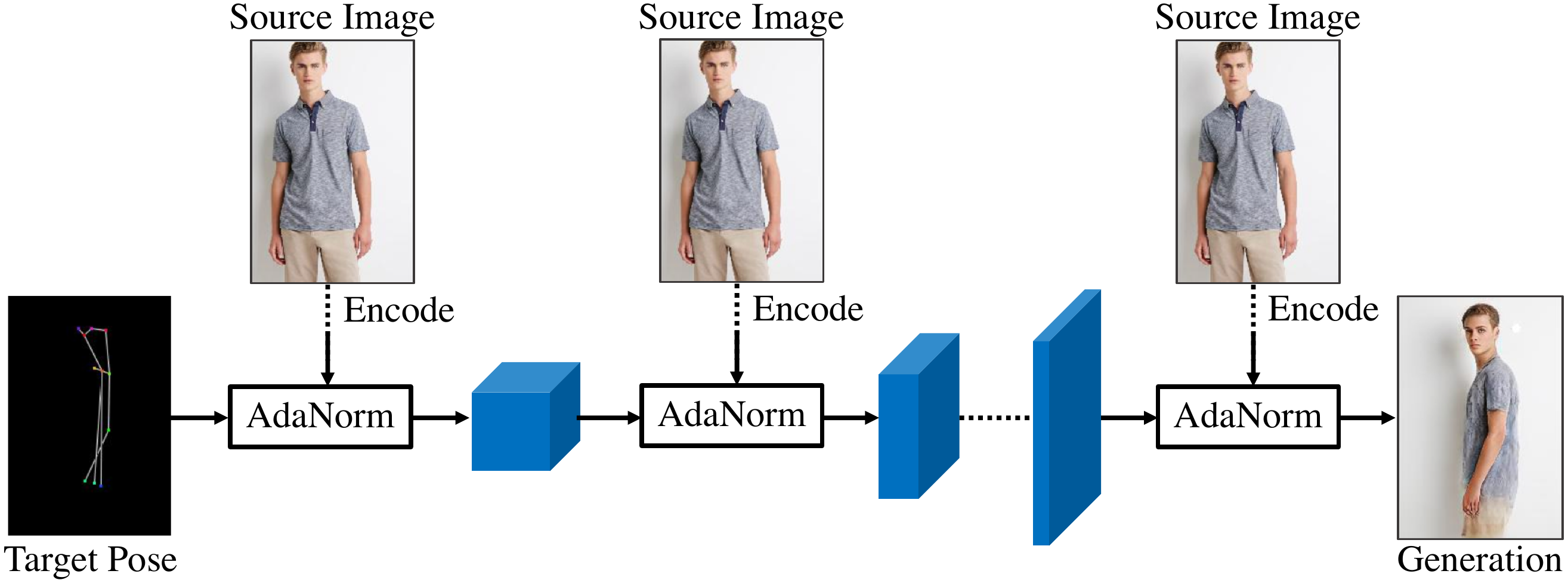}
\caption{A simplified diagram of our APS model. The target pose is coupled with the encoded source appearance through AdaNorm module, progressively. }
\label{fig:adanorm}
\end{figure}

Motivated by the above observations, we investigate a strategy to decouple and re-couple the entanglement factors including \emph{appearance--pose}, \emph{foreground--background}, and \emph{local detail--global structure} in person images. We propose a novel end-to-end person image generation framework consisting of an appearance encoder and an Appearance-aware Pose Stylizer (APS): The appearance encoder learns the appearance representation of the person image; APS, as shown in Fig. \ref{fig:adanorm}, is the image generator which progressively couples the target pose map with the appearance representation, enabling a natural re-coupling of pose and appearance. We additionally adopt an attention mechanism in both encoder and generator to disentangle the foreground and background. In APS, the image is progressively synthesized from small to large scale, thus local details and global structures are fused and preserved in a multi-scale approach..

In summary, the proposed end-to-end framework can effectively decouple the entanglements between appearance, pose, foreground, background, local details, and global structures, and re-couple them in the generator, thus generate high-quality person images in a highly flexible and controllable way. The contributions of this paper are summarized below. 
(1) We propose a novel person image generation framework to make explicit disentanglement of the complex factors in human images.
(2) We propose a new normalization method called adaptive patch normalization. It enables normalization within local regions and is suited to the spatially-dependent generative tasks including person image generation.
(3) We have conducted extensive quantitative and qualitative experiments, and ablation studies to validate the effectiveness of the proposed methods. 

\begin{figure*}[t]
\centering
\includegraphics[width=1\linewidth]{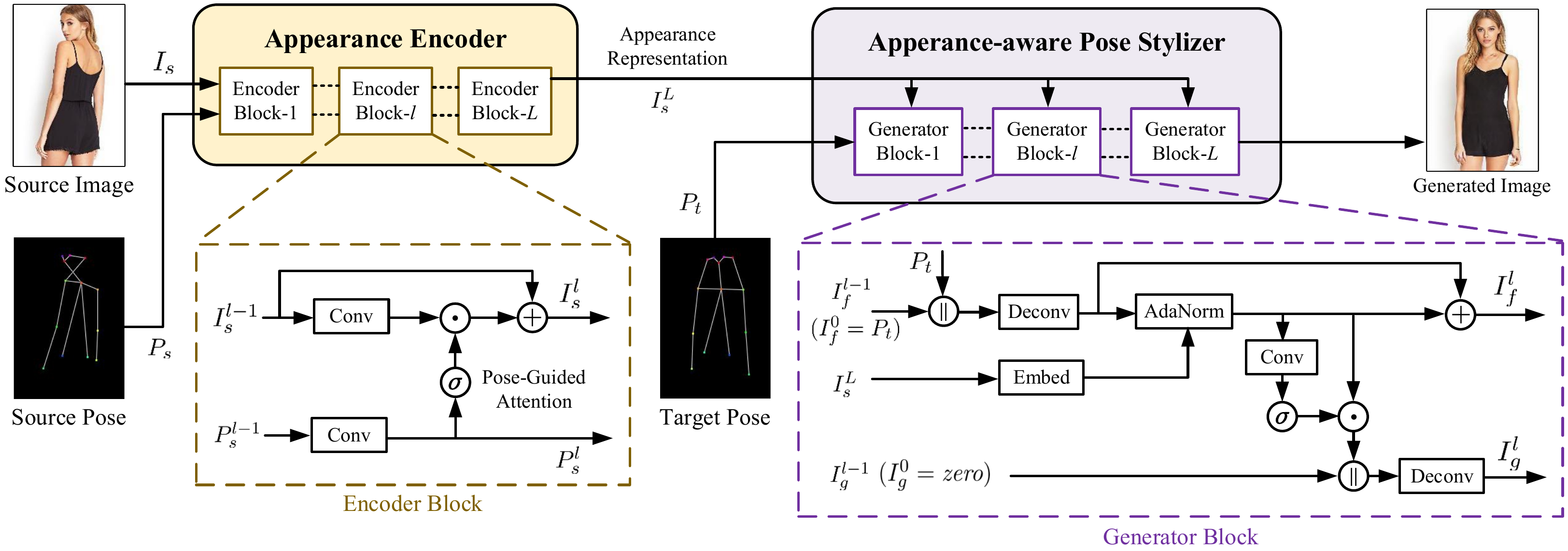}
\caption{The schematic of the proposed person image generation framework. The scheme consists of an appearance encoder which learns the person appearance representation and an Appearance-aware Pose Stylizer which generates the target image. Notations: $\sigma$ sigmoid activation, $\cdot$ element-wise multiplication, $+$ element-wise addition, $||$ channel-wise concatenation. }
\label{fig:framework}
\end{figure*}

\section{Related Work}

Person image generation is very valuable in many real-world applications \cite{wei2018person,chan2019everybody}. Various settings of person image generation have been proposed in the literature. \cite{lassner2017generative} proposed 3D pose representation to generate images of a person with different clothes. \cite{zhao2018multi} generated multi-view cloth images from
a single view cloth image. Several work on virtual try-on \cite{han2018viton,wang2018toward,zanfir2018human,dong2019towards} including FashionGAN \cite{fashiongan} proposed to manipulate the clothes of a given person while maintaining the person identity and pose. 

In this work we focus on the pose-guided person image generation \cite{pg2} which aims at generating images of a person with different poses but with the same clothes and identity. Based on generative adversarial networks (GANs) \cite{gan,conditionalgan}, several efforts \cite{neverova2018dense,Song_2019_CVPR,Zhou_2019_CVPR} have been made towards this goal. More related to this work, \cite{disentangle} proposed to disentangle image foreground, background, and pose into separated representations by the encoder. \cite{patn} proposed to transfer person pose in the encoding phase using an attention-based progressive model \cite{progressivegan}. Both the above methods disentangles image factors in the encoding phase and combines the disentangled representations before decoding, lacking explicit cross-modal re-coupling in the decoding phase. 

Our approach is also related to adaptive normalization-based GANs \cite{stylegan,spade} which progressively transforms a constant vector or a random noise (namely, \emph{content}) into an image using a stack of convolutional layers and normalization layers with learned coefficients (namely, \emph{style}). More recently, \cite{yildirim2019generating} applied StyleGAN \cite{stylegan} to fashion image generation, where the content input is a constant vector and the style input is a combination of clothes and pose information. Different from existing normalization-based generative models, we set pose and appearance information of a person as content and style inputs, respectively. Our approach enables a more natural re-coupling of cross-modal representations, thus it is more effective and efficient in articulated-object generation problems.

\section{Our Approach}

The goal of pose-guided person image generation is to generate a person image $I_g$ which is expected to follow a given person pose $P_s$ while keep the appearance details of a given source person image $I_s$. In this paper, we propose an end-to-end generative framework including an appearance encoder and an  Appearance-aware Pose Stylizer (APS) to address this challenging task.

\subsection{Appearance Encoder}

We use an attention-based appearance encoder to learn the appearance representation of the source image $I_s$. The appearance encoder is built upon a stack of $L$ encoder blocks and the architecture is shown in the left part of Fig. \ref{fig:framework}. The appearance encoder has two network streams, the image stream and pose stream. The two streams take the outputs of two streams in layer $l-1$ as their inputs and output $I_s^{l}$ and $P_s^{l}$, respectively. The input of image stream is source image $I_s$ such that image stream extracts the visual representation of source image. The input of pose stream is source pose $P_s$ such that pose stream learns the structure of source pose to guide the information flow in image stream with attention mechanism \cite{patn}. More specifically, the pose input $P_s^{l}$ from previous layer goes through a convolutional layer and a sigmoid activation layer to obtain foreground attention masks $M_s^l \in (0,1)$ as
\begin{equation}
\label{eq:enc_att}
M_s^l=\sigma\left(\mathrm{conv}_{P}\left(P_s^{l-1}\right)\right)
\end{equation} 
Then, $M_s^l$ masks the image stream using element-wise multiplication as 
\begin{equation}
\label{eq:enc}
    I_s^l = M_s^l \odot \mathrm{conv}_{I}\left( I_s^{l-1} \right)+ I_s^{l-1}
\end{equation}
The residual connection in Eq. \ref{eq:enc} eases the learning of deep generative models in practice. With pose-guided attention, image stream is forced to focus on the foreground of image, i.e., the person.

\subsection{Appearance-aware Pose Stylizer}

Appearance-aware Pose Stylizer (APS) is a novel generator for realistic person image generation. As shown in Fig. \ref{fig:adanorm}, the pose stylizer stylizes the target pose map $P_t$ under guidance of appearance representation, progressively. Similar to appearance encoder, APS consists of $L$ repetitive generator blocks to restore images from small sizes to large sizes. In the following we discuss more details of the generator blocks.

\subsubsection{Generator Block}
In a generator block, there are two network streams including foreground image stream and synthesized image stream, as shown in the right part of Fig. \ref{eq:enc_att}. Foreground image stream $I_f$ generates foreground person images based on pose maps $P_t$ and appearance representations $I_s^L$. Synthesized image stream $I_g$ synthesizes complete images including both foregrounds and backgrounds. In the foreground stream, we adopt adaptive normalization mechanism (AdaNorm) to stylize pose maps based on appearance representations, such that our generator is named as Appearance-aware Pose Stylizer. $\textrm{AdaNorm}(x,y)$ accepts content feature $x$ and style feature $y$
\begin{equation}
    x = \mathrm{deconv}_{f}\left(\mathrm{concat}\left(I_f^{l-1}, P_t\right) \right)
\end{equation}
\begin{equation}
    y = \mathrm{embed}\left( I_s^{L} \right)
\end{equation}
$\mathrm{deconv}$ is a deconvolutional layer, $\mathrm{concat}$ is the channel-wise concatenation operation, $\mathrm{embed}$ is a convolutional layer with a kernel size of 1$\times$1. $I_f^0 = P_t$ such that content $x$ is derived from pose map $P_t$. Style $y$ is derived from appearance representation $I_s^{L}$. Note that the setting of content and style in this work is distinctly different from existing normalization-based generative models such as StyleGAN \cite{stylegan} and SPADE \cite{spade}, in which the content $x$ is generally set as a constant vector or a random vector. In our setting, AdaNorm naturally disentangles pose and appearance of a person as its content and style inputs, leading to a reasonable feature fusion as well as an effective person generation pipeline.

After computing $z = \textrm{AdaNorm}(x, y)$, the foreground stream outputs $I_f^l = z + x$. The foreground stream fuses into the synthesized stream with attention mechanism,
\begin{equation}
   z_{\textrm{Att}} =  \sigma\left(\mathrm{conv}_{f}\left(z\right)\right) \odot z
\end{equation}
The synthesized stream outputs $I_g^l$ as
\begin{equation}
    I_g^l = \mathrm{deconv}_{g}\left(\mathrm{concat}\left(I_g^{l-1}, z_{\textrm{Att}}\right) \right)
\end{equation}
Specifically, $I_g^0$ is feature maps with $\emph{zero}$ values. $I_f^L$ and $I_g^L$ output by the last generator block are concatenated and decoded as the generated image with an 1$\times$1 convolutional layer.

\subsubsection{Adaptive Patch Normalization}
Here we discuss the AdaNorm module used in our method. In existing literature, StyleGAN and SPADE successfully applied adaptive instance normalization (AdaIN) \cite{adain} to progressive generative models. AdaIN is formulated as
\begin{equation}
\label{eq:adain}
\textrm{AdaIN}(x_c, y)= y_c^w \left(\frac{x_c-\beta(x_c)}{\gamma(x_c)}\right)+y_c^b
\end{equation}
where $x$ and $y$ are content and style features respectively. $c$ denotes the channel number, $\beta$ and $\gamma$ denote mean and standard deviation. The weight term $y_c^w$ and bias term $y_c^b$ are embedded from input style feature $y$. In Eq. \ref{eq:adain}, content feature $x$ is first normalized by instance normalization (IN) and then scaled by parameters conditioned on style feature $y$.

\begin{figure}[t]
\centering
\includegraphics[width=1\linewidth]{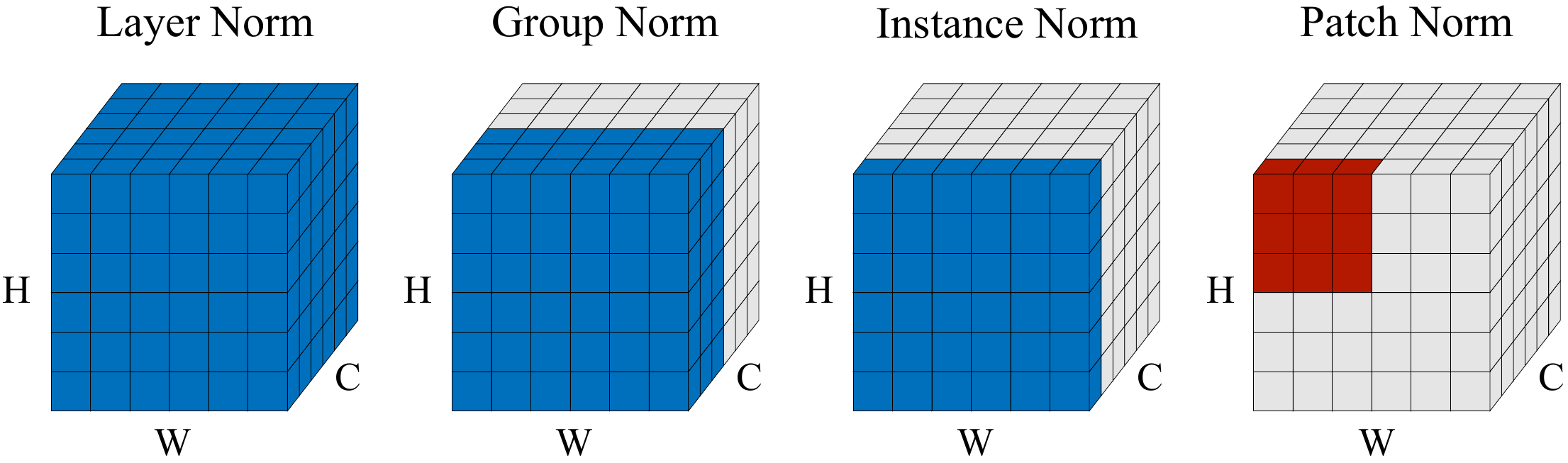}
\caption{Illustrations of normalization methods. Compared with existing methods, patch normalization enables region-specific normalization parameters.}
\label{fig:norms}
\end{figure}

In this paper we develop a new normalization method, i.e., adaptive patch normalization (AdaPN)
\begin{equation}
\label{eq:patchnorm}
\textrm{AdaPN}(x_{c,i,j}, y)= y_{c,i,j}^w \left(\frac{x_{c,i,j}-\beta(x_{c,i,j})}{\gamma(x_{c,i,j})}\right)+y_{c,i,j}^b
\end{equation}
where $i,j$ denote the spatial position of a patch on feature map. The content feature is first normalized by IN and then scaled with region-specific parameters $y_{c,i,j}^w$ and $y_{c,i,j}^b$. \footnote{In implementation of AdaPN, in each generator block we embed $I_s^L$ to latent features and then tile the features into $y^w$ and $y^b$. The sizes of $y^w$ and $y^b$ are the same to the size of corresponding $x$.}
Fig. \ref{fig:norms} illustrates the difference between existing normalization methods. The proposal of patch normalization is motivated by the observation that human body parts within a specific spatial region are relatively fixed among well-cropped person images. For instance, the head usually appears in the top-center area of an image while the legs usually appear in the bottom area of an image. Therefore, it is natural to normalize spatial regions with different factors. Compared with AdaIN, AdaPN induces the generator to learn appearance details of different body parts more effectively. In Section \ref{sec:norm}, we take insights into AdaPN by conducting more empirical studies.

\subsection{Training and Optimization}

The loss function that we use to train our person image generation model is comprised of the adversarial loss $\mathcal{L}_{\textrm{GAN}}$, the reconstruction loss ${L_1}$, and the perceptual loss $\mathcal{L}_{\textrm{per}}$ as follow:
\begin{equation}
\label{eq:total_loss}
\mathcal{L} = \arg\underset{G}{\min}\, \underset{D}{\max}\,  \alpha \mathcal{L}_{\textrm{GAN}} + \lambda_1 \mathcal{L}_{L1} + \lambda_2 \mathcal{L}_{\textrm{per}}
\end{equation}
In Eq. \ref{eq:total_loss}, ${L_1}$ loss $\mathcal{L}_{L1} = \| I_g - I_t \|_1 $ where $I_g$ is the generated image and $I_t$ is the ground-truth target image. The perceptual loss \cite{perceptual} computes ${L_1}$ distance over feature maps. The adversarial loss $\mathcal{L}_{\textrm{GAN}}$ consists of the appearance-consistency term and the pose-consistency term, formulated as
\begin{equation}
\begin{split}
\mathcal{L}_{\textrm{GAN}} = & \mathbb{E} [~\underbrace{\log D_a(I_s, I_t) + \log (1- D_a(I_s, I_g))}_{\text{appearance-consistency term}} \\
& + \underbrace{\log D_p(P_t, I_t) + \log (1- D_p(P_t, I_g))}_{\text{pose-consistency term}}~]
\end{split}
\end{equation}
where $D_a$ and $D_p$ are discriminators. $(I_s, I_t) \sim \mathbb{I}_{\text{real}}, I_g \sim \mathbb{I}_{\text{fake}}, P_t \sim \mathbb{P}$. $\mathbb{I}_{\text{real}}$, $\mathbb{I}_{\text{fake}}$, $\mathbb{P}$ are the distribution of real images, fake images, and person poses, respectively. 

\input{fashion_baseline.tex}

\noindent
\textbf{Network architectures.}
For both encoder and generator, we adopt a total block number $L=4$ on Market-1501 dataset and $L=5$ on DeepFashion dataset, respectively. The first layer of encoder and the last layer of generator has 64 channels. The number of channels in every block is doubled/halved in encoder/generator until a maximum of 512. The size of feature map in every block is halved/doubled in encoder/generator using stride-2 convolutions/deconvolutions. The appearance representation $I_s^L$ has a shape of $512 \times \frac{H}{2^L} \times \frac{W}{2^L}$, where $H$ and $W$ is height and width of the input image. Every AdaNorm module is built up with AdaPN-Conv-AdaPN in practice for establishing a deeper model.The standard/adaptive normalization layer and ReLU are applied after every convolutional or deconvolutional layer. 

\noindent
\textbf{Training details.}
We implement our model on the PyTorch framework \cite{pytorch}. The model is trained with an Adam optimizer \cite{adam} for 800 epochs. The initial learning rate is 0.0002 and it linearly decays to 0 from 400 epochs to 800 epochs. Following \cite{patn}, the loss weights ($\alpha$,$\lambda_1$,$\lambda_2$) are set as (5, 10, 10) on Market-1501 and (5, 1, 1) on DeepFashion. In training Market-1501, we additionally apply Dropout \cite{dropout} with a rate of 0.5 after every generator block in case of overfitting. 

\section{Experiments}

\subsection{Experimental Setups}

\noindent
\textbf{Datasets.}
We conduct experiments on two benchmark person image generation datasets including Market-1501 \cite{market} and DeepFashion (\textit{In-shop Clothes Retrieval Benchmark}) \cite{deepfashion}. Market-1501 is a challenging person re-identification dataset which contains 32,668 images of 1,501 person identities. The images in Market-1501 are low-resolution (128$\times$64 pixels) and the person pose, viewpoint, illumination, and background vary largely. DeepFashion contains 52,712 in-shop clothes images (256$\times$256 pixels). We adopt OpenPose \cite{hpe} as our pose keypoints detector. By following the settings in \cite{patn}, for Market-1501, we collect 263,632 training pairs and 12,000 testing pairs. For DeepFashion, we collect 101,966 pairs for training and 8,570 pairs for testing. Each pair is composed of two images of the same identity but different poses. The person identities in training sets do not overlap with those in testing sets.

\noindent
\textbf{Evaluation metrics.} In this work we use Structural Similarity (\textit{SSIM}) \cite{ssim} to measure the structure similarity between images, i.e., the appearance-consistency. We use the Inception Score (\textit{IS}) \cite{is} to measure the image quality. Following PG2 \cite{pg2} we adopt their masked versions \textit{mask-SSIM} and \textit{IS} to evaluate the image foreground only via masking out the background, since no background information is provided for person generation models. In addition, we use Percentage of Correct Keypoints (\textit{PCKh}) \cite{pckh} to measure the pose joints alignment, i.e., the pose-consistency.

\begin{table*}[h]
    \newcommand{\width}{1}
	\footnotesize
	\centering
	\begin{tabular}{|l|p{\width cm}<{\centering} 
	p{\width cm}<{\centering} 
	p{1.55 cm}<{\centering} 
	p{1.2 cm}<{\centering} 
	p{\width cm}<{\centering}
	|p{\width cm}<{\centering} 
	p{\width cm}<{\centering} 
	p{\width cm}<{\centering}|}
		\hline
		\multirow{2}{*}{\textbf{Model}} & \multicolumn{5}{c|}{\textbf{Market-1501}}           & \multicolumn{3}{c|}{\textbf{DeepFashion}} \\ \cline{2-9} 
		& SSIM  & IS    & mask-SSIM & mask-IS & PCKh & SSIM      & IS       & PCKh   \\ \hline
		\textit{Real Data}  & 1.000 & 3.890  & 1.000  & 3.706  & 1.00   & 1.000  & 4.053    & 1.00        \\  \hline
		PG2 \cite{pg2} & 0.261 & \textbf{3.495} & 0.782     & 3.367   & 0.73  & \underline{\textbf{0.773}}    & 3.163  & 0.89   \\
		Disentangled \cite{disentangle}    & 0.099 & \underline{\textbf{3.483}} & 0.614     & 3.491   & -     & 0.614    & 3.228    & -     \\
		VUNet \cite{vunet}          & 0.266 & 2.965 & 0.793 & 3.549 & 0.92 & 0.763 & \textbf{3.440} & 0.93 \\
		Deform \cite{deform}           & 0.291 & 3.230          & 0.807     & 3.502   & \textbf{0.94}  & 0.760    & \underline{\textbf{3.362}}      & 0.94  \\  
		PATN \cite{patn}             & \underline{\textbf{0.311}}  & 3.323  & \textbf{0.811} & \textbf{3.773}   & \textbf{0.94} & \underline{\textbf{0.773}}  & 3.209 & \textbf{0.96} \\  \hline
		Ours  & \textbf{0.312} & 3.132 & \underline{\textbf{0.808}} & \underline{\textbf{3.729}} & \textbf{0.94} & \textbf{0.775} & 3.295 & \textbf{0.96} \\  \hline
	\end{tabular}
	\caption{Quantitative comparison of our proposed method to the state-of-the-art methods on Market-1501 and DeepFashion. The \textbf{best} and the \underline{\textbf{second-best}} performances are highlighted (Higher is better for all reported metrics).}
	\label{table:quan}
\end{table*}

\subsection{Results}

\noindent
\textbf{Qualitative comparison.}
Fig. \ref{fig:fashion} shows a qualitative comparison of state-of-the-art person image generation methods on DeepFashion. The source images, pose maps, target images, and the generations are shown from the left to the right, respectively. We compare our method with Variational U-Net (VUNet) \cite{vunet}, Deformable GANs (Deform) \cite{deform}, and Pose-Attentional Transfer Network (PATN) \cite{patn}. The example images shown in Fig. \ref{fig:fashion} vary in poses, scales, and colors and types of clothes. Our method shows appealing results in the following aspects: 
(1) \emph{Realistic generations}: The generated images show natural facial details, body postures, clothing collocations, and skin appearances;
\noindent
(2) \emph{Clothing-consistency}: Clothes colors and styles are consistent with those in source images;
\noindent
(3) \emph{Identity-consistency}: Person identity details including facial appearances, body figures, and skin colors are well maintained in generated images;
\noindent
(4) \emph{Pose-consistency}: Poses of generated persons well follow the conditioned poses.

\begin{figure}[t]
\centering
\small

\includegraphics[width=0.115\linewidth]{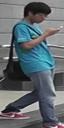}
\includegraphics[width=0.115\linewidth]{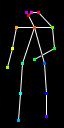}
\includegraphics[width=0.115\linewidth]{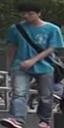}
\includegraphics[width=0.115\linewidth]{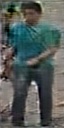}
\includegraphics[width=0.115\linewidth]{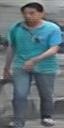}
\includegraphics[width=0.115\linewidth]{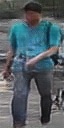}
\includegraphics[width=0.115\linewidth]{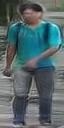}
\includegraphics[width=0.115\linewidth]{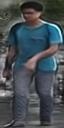}

\vspace{0.02cm}

\stackunder[3pt]{\includegraphics[width=0.115\linewidth]{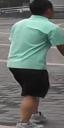}}{\small Source}
\stackunder[3pt]{\includegraphics[width=0.115\linewidth]{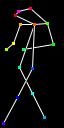}}{\small Pose}
\stackunder[3pt]{\includegraphics[width=0.115\linewidth]{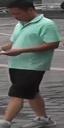}}{\small Target}
\stackunder[3pt]{\includegraphics[width=0.115\linewidth]{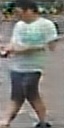}}{\small PG2}
\stackunder[3pt]{\includegraphics[width=0.115\linewidth]{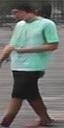}}{\small VUNet}
\stackunder[3pt]{\includegraphics[width=0.115\linewidth]{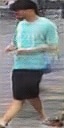}}{\small Deform}
\stackunder[3pt]{\includegraphics[width=0.115\linewidth]{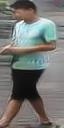}}{\small PATN}
\stackunder[3pt]{\includegraphics[width=0.115\linewidth]{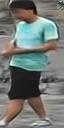}}{\small Ours}
\caption{Qualitative comparison of existing pose-guided person generation methods on Market-1501. 
}
\label{fig:market}
\end{figure}

Fig. \ref{fig:market} shows qualitative results on Market-1501. Although the images of this dataset are low-resolution and their visual details are somewhat blurry (as shown in Source and Target columns), we observe that our method shows a good performance in comparison with the other methods considering the sharpness of generated images. 

\begin{table*}[t]
    \newcommand{\width}{1}
	\footnotesize
	\centering
	\begin{tabular}{|l|p{\width cm}<{\centering} 
	p{\width cm}<{\centering} 
	p{\width cm}<{\centering} 
	p{1.8 cm}<{\centering} 
	p{1.35 cm}<{\centering}
	p{1.4 cm}<{\centering} 
	|p{\width cm}<{\centering} 
	p{\width cm}<{\centering}
	p{\width cm}<{\centering}|}
		\hline
		\multirow{2}{*}{\textbf{Model}} & \multicolumn{6}{c|}{\textbf{Market-1501}}           & \multicolumn{3}{c|}{\textbf{DeepFashion}} \\ \cline{2-10} 
		& SSIM $\uparrow$  & IS $\uparrow$ & L1 $\downarrow$   & mask-SSIM $\uparrow$ & mask-IS $\uparrow$ & mask-L1 $\downarrow$ & SSIM  $\uparrow$    & IS  $\uparrow$     & L1 $\downarrow$  \\ \hline
		Conv enc+Deconv dec  & 0.205 & 3.141 & 0.332 & 0.750 & 3.507 & 0.104 & 0.758 &  3.373 & 0.106 \\ 
		Conv enc+APS dec  & 0.301 & 2.985 & 0.286 & 0.804 & 3.661 & 0.080 & 0.766 & 3.335 & 0.101 \\ 
		PATN enc+Deconv dec  & 0.218 & \textbf{3.193} & 0.341   & 0.751   & 3.447  & 0.101    & 0.760  & 3.334 & 0.104   \\ \hline
		
		StyleGAN     & 0.251 & 2.987 & 0.312   & 0.777    & 3.737    & 0.093   & 0.766 & \textbf{3.393} & 0.101    \\
		APS w/ AdaIN        & 0.297 & 3.094 & 0.293 & 0.800 & \textbf{3.755} & 0.081 & 0.763 & 3.275 & 0.102\\
		APS w/o attention   & 0.291  & 2.879  & 0.292 & 0.799  & 3.653 & 0.082  & 0.764 & 3.305 & 0.100 \\   
		APS w/o decoding $I_f^L$ & 0.303 & 2.993 & 0.285 & 0.806 & 3.622 & \textbf{0.079} & 0.768 & 3.374 & 0.098  \\ \hline  
		
		Full model & \textbf{0.312} & 3.132 & \textbf{0.281} & \textbf{0.808} & 3.729 & \textbf{0.079} & \textbf{0.775} & 3.295 & \textbf{0.097} \\  \hline
	\end{tabular}
	\caption{Ablation studies on our person image generation model for evaluating the efficacy of different components. $\uparrow$ denotes higher is better and $\downarrow$ denotes lower is better. }
	\label{table:ablation}
\end{table*}

\noindent
\textbf{Quantitative comparison.} Table \ref{table:quan} quantitatively evaluate the person generation methods under a series of metrics. Our method shows a competitive performance compared with the existing methods. It achieves the best PCKh on both datasets, indicating that the generations have a good consistency with the conditioned poses. We attribute it to the natural disentanglement of pose and appearance in our APS model. On Market-1501, our model performs well on SSIM, mask-SSIM, and mask-IS. On DeepFashion, our model performs well on SSIM. However, it is relatively worse on IS. We conjecture it is because the model well restores the visual appearance details, while, a certain level of overfitting may hurt the realness performance. 

\subsection{Ablation Study}

We conduct ablation study on our person generation model to evaluate the efficacy of different components proposed in this paper. The first part of Table \ref{table:ablation} evaluates different encoders and decoders. With the same encoders (Conv encoder or PATN encoder), our APS decoder shows large improvements over Deconv decoder. The second part of Table \ref{table:ablation} evaluates different components of APS generator.  In StyleGAN \cite{stylegan}, the content input is a constant vector and the style input contains both pose and appearance representation. APS shows significant improvements over StyleGAN, demonstrating that the distinct disentanglement of pose and appearance in generator can benefit the performance of person image generation. APS w/ AdaIN replaces the AdaPN with AdaIN in APS, and the results show that our proposed AdaPN is more suited to spatially-dependent generative tasks. APS w/o attention removes all the attention modules in encoder and decoder, and the results indicate that attention mechanism can slightly help the APS model. Decoding both $I_f^L$ and $I_g^L$ is better than only decoding the synthesized stream $I_g^L$.

\begin{figure}[t]
\centering
\includegraphics[width=0.49\linewidth]{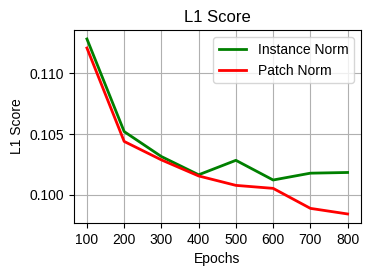}
\hspace{0.1cm}
\includegraphics[width=0.47\linewidth]{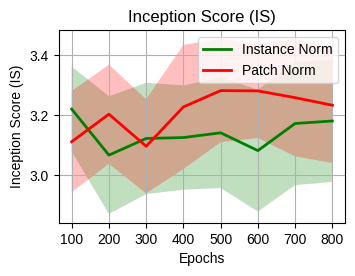}
\caption{A comparison of AdaIN and AdaPN. We show L1 score (lower is better) and Inception Score (higher is better) vs. training epochs on testing set of DeepFashion. AdaPN is better than AdaIN under both metrics.}
\label{fig:quan_norm}
\end{figure}

\subsection{Study on Adaptive Normalization}

Adaptive normalization is the core of our APS model. Fig. \ref{fig:quan_norm} compares AdaIN and AdaPN on testing set of DeepFashion, vs., the training epochs. Before 400 epochs, AdaIN and AdaPN shows similar L1 scores. After 400 epochs, L1 scores of AdaIN does not decrease anymore, while, L1 scores of AdaPN is decreasing continuously until the end of training. It reveals that AdaPN has a larger model capacity than AdaIN in learning, thus to better reconstruct the high resolution images. AdaPN also shows better Inception Scores on most of the epochs, demonstrating its superiority in person generation model. 

\begin{figure}[t]
\centering
\includegraphics[width=1\linewidth]{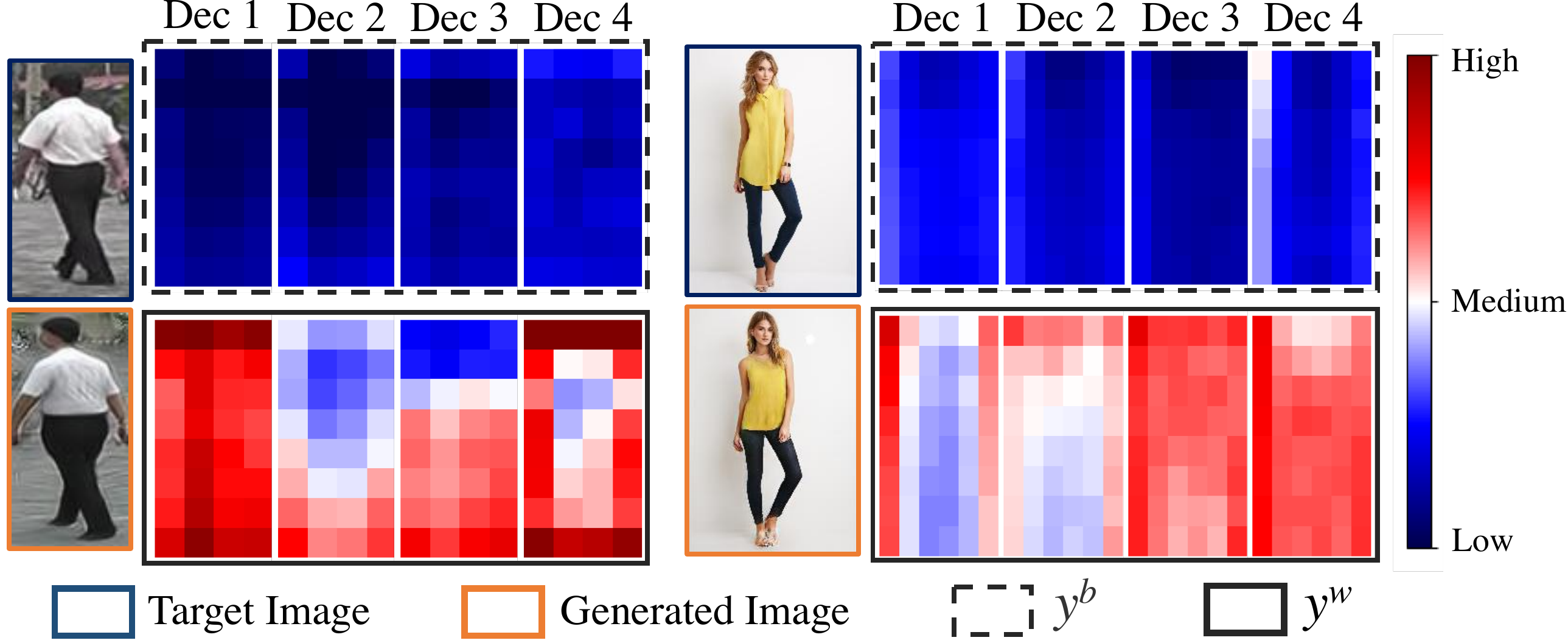}
\caption{Visualizations of AdaPN statistics, including bias $y^b$ and scaling factor $y^w$, in individual generator blocks.}
\label{fig:viz_norm}
\end{figure}

In Fig. \ref{fig:viz_norm} we visualize the AdaPN statistics of a fully trained model. The statistics of individual generator blocks, including bias $y^b$ and scaling factor $y^w$, are shown at right. We notice that the biases of different positions are similar within a layer. Conversely, the scaling factors of different positions vary largely, suggesting that the local-version scaling is valuable in normalization-based person generation, while, the local-version translation is not much necessary. Intuitively, the scaling operation is related with the input variables. Compared with the bias term, the scaling term contributes more to variations within the output. This leads to the necessity of locally sensitive scaling factors. 

\label{sec:norm}

\section{Conclusion}
In this paper, we have presented a novel framework for generating realistic person images. The framework decouples the image factors by an attention-based appearance encoder and re-couples the image factors by an APS generator. It is effective, controllable, and flexible since it makes explicit disentanglement of the complex factors in human images. We have also proposed AdaPN which enables local-specific normalization for spatially-dependent generative tasks. Extensive experiments on benchmark datasets have validated the effectiveness of our approach over existing methods.

\bibliographystyle{ijcai}
{\small
\bibliography{main}}

\end{document}

%% file: fashion_baseline.tex
\newcommand{\scale}{0.0655}
\newcommand{\hscale}{0.3}
\newcommand{\vscale}{0.05}
\begin{figure*}[t]
\centering
\small

\includegraphics[width=\scale \linewidth]{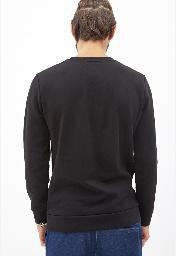}
\includegraphics[width=\scale \linewidth]{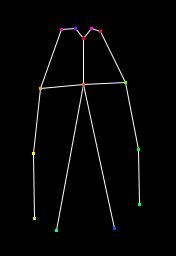}
\includegraphics[width=\scale \linewidth]{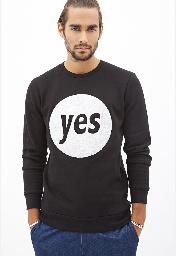}
\includegraphics[width=\scale \linewidth]{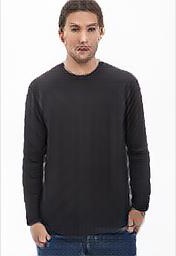}
\includegraphics[width=\scale \linewidth]{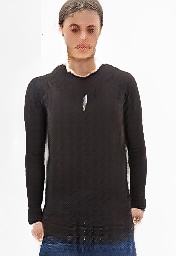}
\includegraphics[width=\scale \linewidth]{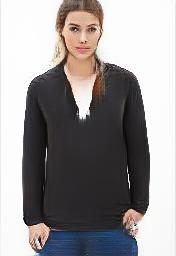}
\includegraphics[width=\scale \linewidth]{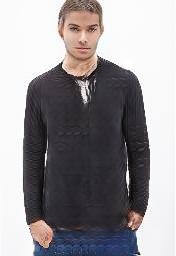}
\hspace{\hscale cm}
\includegraphics[width=\scale \linewidth]{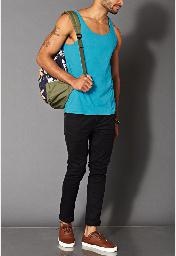}
\includegraphics[width=\scale \linewidth]{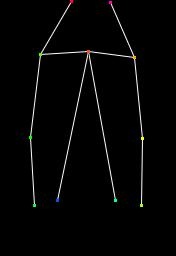}
\includegraphics[width=\scale \linewidth]{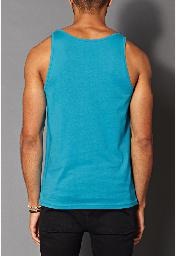}
\includegraphics[width=\scale \linewidth]{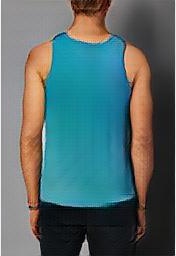}
\includegraphics[width=\scale \linewidth]{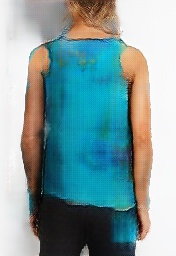}
\includegraphics[width=\scale \linewidth]{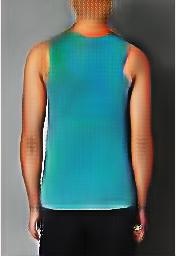}
\includegraphics[width=\scale \linewidth]{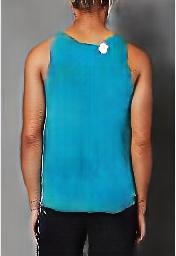}
\vspace{\vscale cm}

\includegraphics[width=\scale \linewidth]{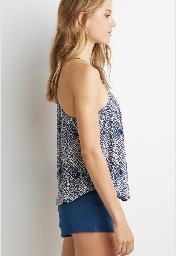}
\includegraphics[width=\scale \linewidth]{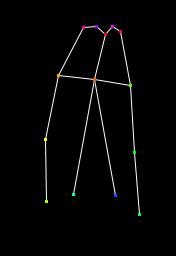}
\includegraphics[width=\scale \linewidth]{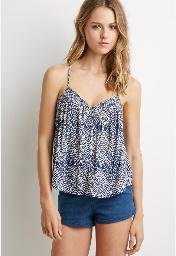}
\includegraphics[width=\scale \linewidth]{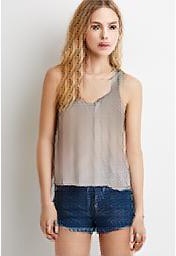}
\includegraphics[width=\scale \linewidth]{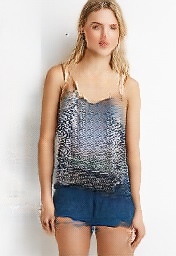}
\includegraphics[width=\scale \linewidth]{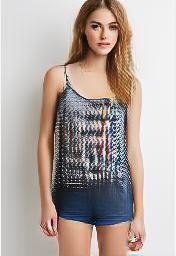}
\includegraphics[width=\scale \linewidth]{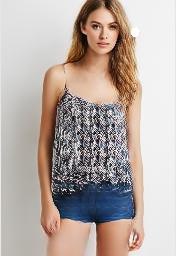}
\hspace{\hscale cm}
\includegraphics[width=\scale \linewidth]{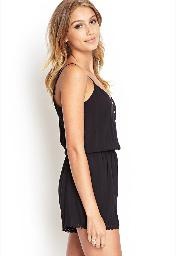}
\includegraphics[width=\scale \linewidth]{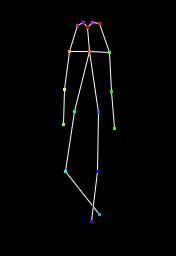}
\includegraphics[width=\scale \linewidth]{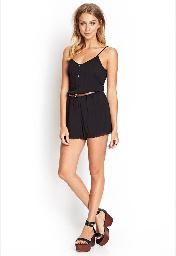}
\includegraphics[width=\scale \linewidth]{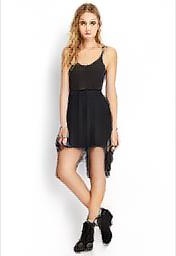}
\includegraphics[width=\scale \linewidth]{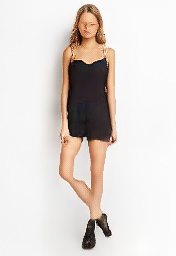}
\includegraphics[width=\scale \linewidth]{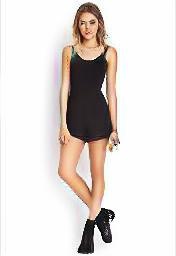}
\includegraphics[width=\scale \linewidth]{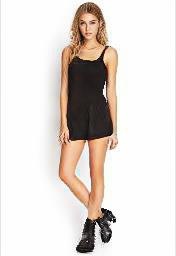}
\vspace{\vscale cm}

\includegraphics[width=\scale \linewidth]{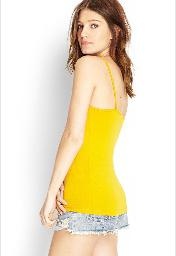}
\includegraphics[width=\scale \linewidth]{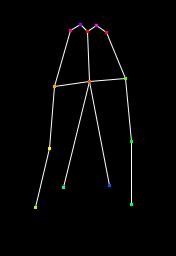}
\includegraphics[width=\scale \linewidth]{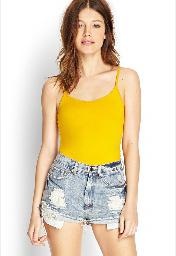}
\includegraphics[width=\scale \linewidth]{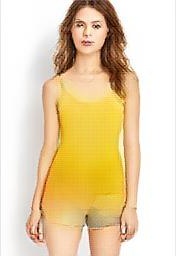}
\includegraphics[width=\scale \linewidth]{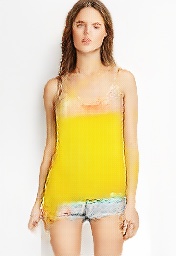}
\includegraphics[width=\scale \linewidth]{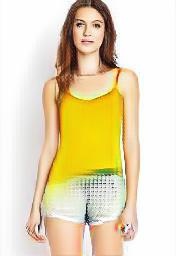}
\includegraphics[width=\scale \linewidth]{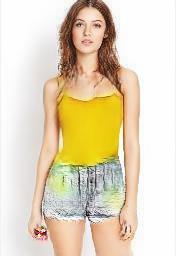}
\hspace{\hscale cm}
\includegraphics[width=\scale \linewidth]{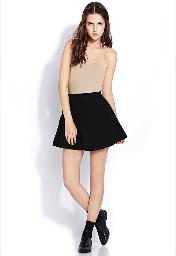}
\includegraphics[width=\scale \linewidth]{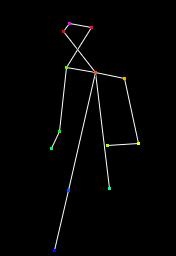}
\includegraphics[width=\scale \linewidth]{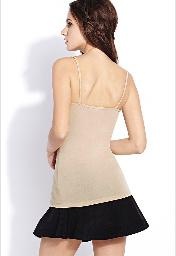}
\includegraphics[width=\scale \linewidth]{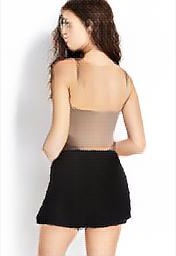}
\includegraphics[width=\scale \linewidth]{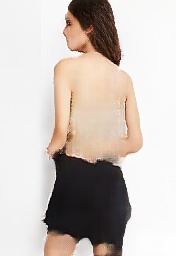}
\includegraphics[width=\scale \linewidth]{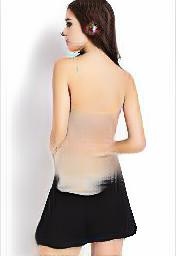}
\includegraphics[width=\scale \linewidth]{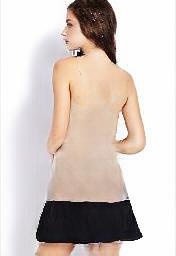}
\vspace{\vscale cm}

\stackunder[3pt]{\includegraphics[width=\scale \linewidth]{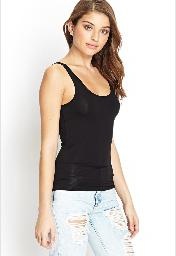}}{\small Source}
\stackunder[3pt]{\includegraphics[width=\scale\linewidth]{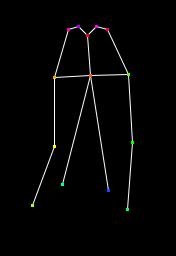}}{\small Pose}
\stackunder[3pt]{\includegraphics[width=\scale\linewidth]{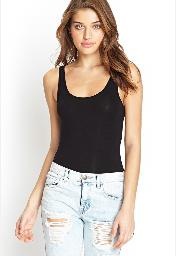}}{\small Target}
\stackunder[3pt]{\includegraphics[width=\scale\linewidth]{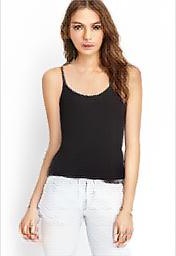}}{\small VUNet}
\stackunder[3pt]{\includegraphics[width=\scale\linewidth]{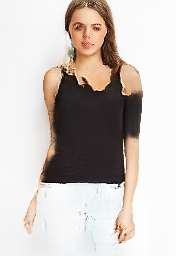}}{\small Deform}
\stackunder[3pt]{\includegraphics[width=\scale\linewidth]{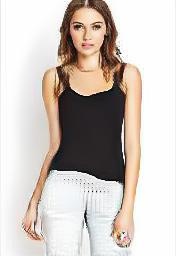}}{\small PATN}
\stackunder[3pt]{\includegraphics[width=\scale\linewidth]{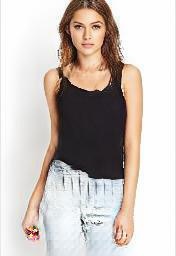}}{\small Ours}
\hspace{\hscale cm}
\stackunder[3pt]{\includegraphics[width=\scale \linewidth]{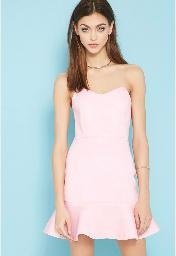}}{\small Source}
\stackunder[3pt]{\includegraphics[width=\scale\linewidth]{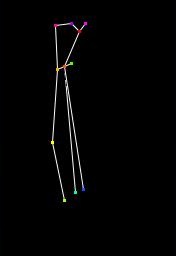}}{\small Pose}
\stackunder[3pt]{\includegraphics[width=\scale\linewidth]{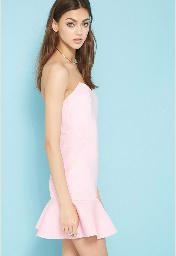}}{\small Target}
\stackunder[3pt]{\includegraphics[width=\scale\linewidth]{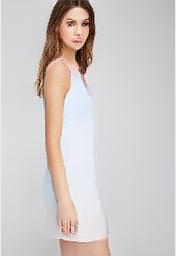}}{\small VUNet}
\stackunder[3pt]{\includegraphics[width=\scale\linewidth]{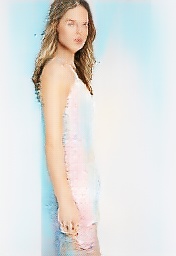}}{\small Deform}
\stackunder[3pt]{\includegraphics[width=\scale\linewidth]{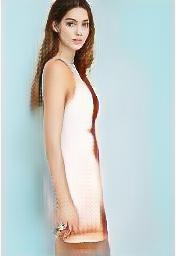}}{\small PATN}
\stackunder[3pt]{\includegraphics[width=\scale\linewidth]{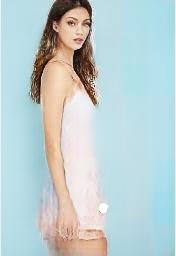}}{\small Ours}
\caption{Qualitative comparison of existing pose-guided person generation methods on DeepFashion. Please zoom in for details.
}
\label{fig:fashion}
\end{figure*}